\documentclass[sigconf]{acmart}

\usepackage{booktabs} % For formal tables
\usepackage{enumitem}
\usepackage{algorithm}% http://ctan.org/pkg/algorithms
\usepackage{algpseudocode}% http://ctan.org/pkg/algorithmicx
\usepackage{xcolor}
\usepackage{tikz}
\usepackage{balance}
\usetikzlibrary{shapes}

\newcommand\shout[1]{#1}

\setcopyright{acmcopyright}
\begin{document}
\title{Graph Coloring: Comparing Cluster Graphs to Factor Graphs}
%\titlenote{Produces the permission block, and copyright information}
%\subtitle{Extended Abstract}
%\subtitlenote{The full version of the author's guide is available as \texttt{acmart.pdf} document}

\copyrightyear{2017}
\acmYear{2017}
\setcopyright{acmcopyright}
\acmConference{SAWACMMM'17}{October 23, 2017}{Mountain View, CA, USA}\acmPrice{15.00}\acmDOI{10.1145/3132711.3132717}
\acmISBN{978-1-4503-5505-6/17/10}

\author{Simon Streicher}
%\authornote{Dr.~Trovato insisted his name be first.}
\affiliation{%
  \institution{ %Electrical and
  Electrical and Electronic Engineering\\Stellenbosch University}
  %\streetaddress{P.O. Box 1212}
  \city{Stellenbosch}
  \state{South Africa}
}
\email{sfstreicher@gmail.com}

\author{Johan du Preez}
%\authornote{The secretary disavows any knowledge of this author's actions.}
\affiliation{%
  \institution{ %Electrical and
  Electrical and Electronic Engineering\\Stellenbosch University}
  %\streetaddress{P.O. Box 1212}
  \city{Stellenbosch}
  \state{South Africa}
}
\email{dupreez@sun.ac.za}

\begin{abstract}
        We present a means of formulating and solving graph coloring
        problems with probabilistic graphical models. In contrast to
        the prevalent literature that uses factor graphs for this
        purpose, we instead approach it from a cluster graph
        perspective. Since there seems to be a lack of algorithms to
        automatically construct valid cluster graphs, we provide such
        an algorithm (termed LTRIP).  Our experiments indicate a
        significant advantage for preferring cluster graphs over
        factor graphs, both in terms of accuracy as well as
        computational efficiency.
\end{abstract}


 \begin{CCSXML}
<ccs2012> <concept>
<concept_id>10002950.10003648.10003649</concept_id>
<concept_desc>Mathematics of computing~Probabilistic
representations</concept_desc>
<concept_significance>500</concept_significance> </concept> <concept>
<concept_id>10002950.10003648.10003670</concept_id>
<concept_desc>Mathematics of computing~Probabilistic reasoning
algorithms</concept_desc>
<concept_significance>500</concept_significance> </concept> <concept>
<concept_id>10002950.10003648.10003649.10003652</concept_id>
<concept_desc>Mathematics of computing~Factor graphs</concept_desc>
<concept_significance>300</concept_significance> </concept> </ccs2012>
\end{CCSXML}

\ccsdesc[500]{Mathematics of computing~Probabilistic representations}
\ccsdesc[300]{Mathematics of computing~Factor graphs}
\ccsdesc[300]{Mathematics of computing~Probabilistic reasoning algorithms}

%\keywords{ACM proceedings, \LaTeX, text tagging}

\maketitle

\section{Introduction}

Due to its learning, inference, and pattern-recognition abilities,
machine learning techniques such as neural networks, probabilistic
graphical models (PGM) and other inference-based algorithms have
become quite popular in artificial intelligence research. PGMs can
easily express and solve intricate problems with many dependencies,
making it a good match for problems such as graph coloring. The PGM
process is similar to aspects of human reasoning, such as the process
of expressing a problem by using logic and observation, and applying
inference to find a reasonable conclusion. With PGMs it is often
possible to express and solve a problem from easily formulated
relationships and observations, without the need to derive complex
inverse relationships. This can be an aid to problems with many
inter-dependencies that cannot be separated into independent parts to
be approached individually and sequentially.

Although the \emph{cluster} graph topology is well established in the
PGM literature ~\cite{Koller2009}, the overwhelmingly dominant
topology encountered in literature is the \emph{factor} graph. We
speculate that this is at least partially due to the absence of algorithms
to \emph{automatically} construct valid cluster graphs, whereas factor
graphs are trivial to construct. To address this we detail a general
purpose construction algorithm termed LTRIP (Layered Trees Running
Intersection Property). We have been covertly experimenting with this
algorithm for a number of years~\cite{myprasapaper, daniekphd}.

\shout{The graph coloring problem} originated from the literal
coloring of planar maps. It started with the four color map theorem,
first noted by Francis Guthrie in 1852. He conjectured that four
colors are sufficient to color neighboring counties differently for
any planar map. It was ultimately proven by Kenneth Appel and Wolfgang
Haken in 1976 and is notable for being the first major mathematical
theorem with a computer-assisted proof. In general, the graph coloring
problem deals with the labeling of nodes in an undirected graph such
that adjacent nodes do not have the same label. The problem is core to
a number of real world applications, such as scheduling timetables for
university subjects or sporting events, assigning taxis to customers,
and assigning computer programming variables to computer
registers~\cite{lewis2015guide,timetables,briggs1992register}. As
graphical models got popular, message passing provided an exciting new
approach to solving graph coloring and (the closely related)
constraint satisfaction
problems~\cite{moon2006multiple,kroc2009counting}. For constraint
satisfaction the survey propagation message passing technique seems to
be particularly
effective~\cite{braunstein2005survey,maneva2004survey,kroc2012sprevisited,knuth2011art4b}.
These techniques are primarily based on the factor graph PGM topology.

The work reported here forms part of a larger project aimed at
developing an efficient alternative for the above message passing
solutions to graph coloring. Cluster graphs and their efficient
configuration are important in that work, hence our interest in those
aspects here. Although we do also provide basic formulations for
modeling graph coloring problems with PGMs, this is not the primary
focus of the current paper, but instead only serves as a vehicle for
comparing topologies.

This paper is structured as follows. Section~\ref{sec:graph_coloring}
shows how the constraints of a graph coloring problem can be
represented as ``factors''. Furthermore, it is shown how these factors
are linked up into graph structures on which inference can be
applied. Section~\ref{sec:topology} discusses the factor graph and
cluster graph topologies, as well as algorithms for automatically
configuring them. The former is trivial, for the latter we provide the
LTRIP algorithm in Section~\ref{sec:ltrip}. Section~\ref{sec:sudoku}
then integrates these ideas by expressing the well known Sudoku puzzle
(an instance of a graph coloring problem) as a PGM. The experiments in
Section~\ref{sec:experiments} shows that, especially for complex
cases, the cluster graph approach is simultaneously faster and more
accurate. The last two sections consider possible future exploration
and final conclusions.

\section{Graph coloring with PGMs} \label{sec:graph_coloring}

This section provides a brief overview of graph coloring and PGMs,
along with techniques of how to formulate a graph coloring problem as
a PGM. We also explore the four color map theorem and illustrate
through an example how to solve these and similar problems.

\subsection{A general description of graph coloring problems}
Graph coloring problems are NP-complete -- easily defined and verified,
but can be difficult to invert and solve. The problem is of
significant importance as it is used in a variety of combinatorial and
scheduling problems.

The general graph coloring problem deals with attaching labels (or
``colors'') to nodes in an undirected graph, such that (a) no two
nodes connected by an edge may have the same label, and (b) the number
of different labels that may be used is minimized. Our focus is mostly
on the actual labeling of a graph.

\begin{figure}[h]
  \includegraphics[width=\columnwidth]{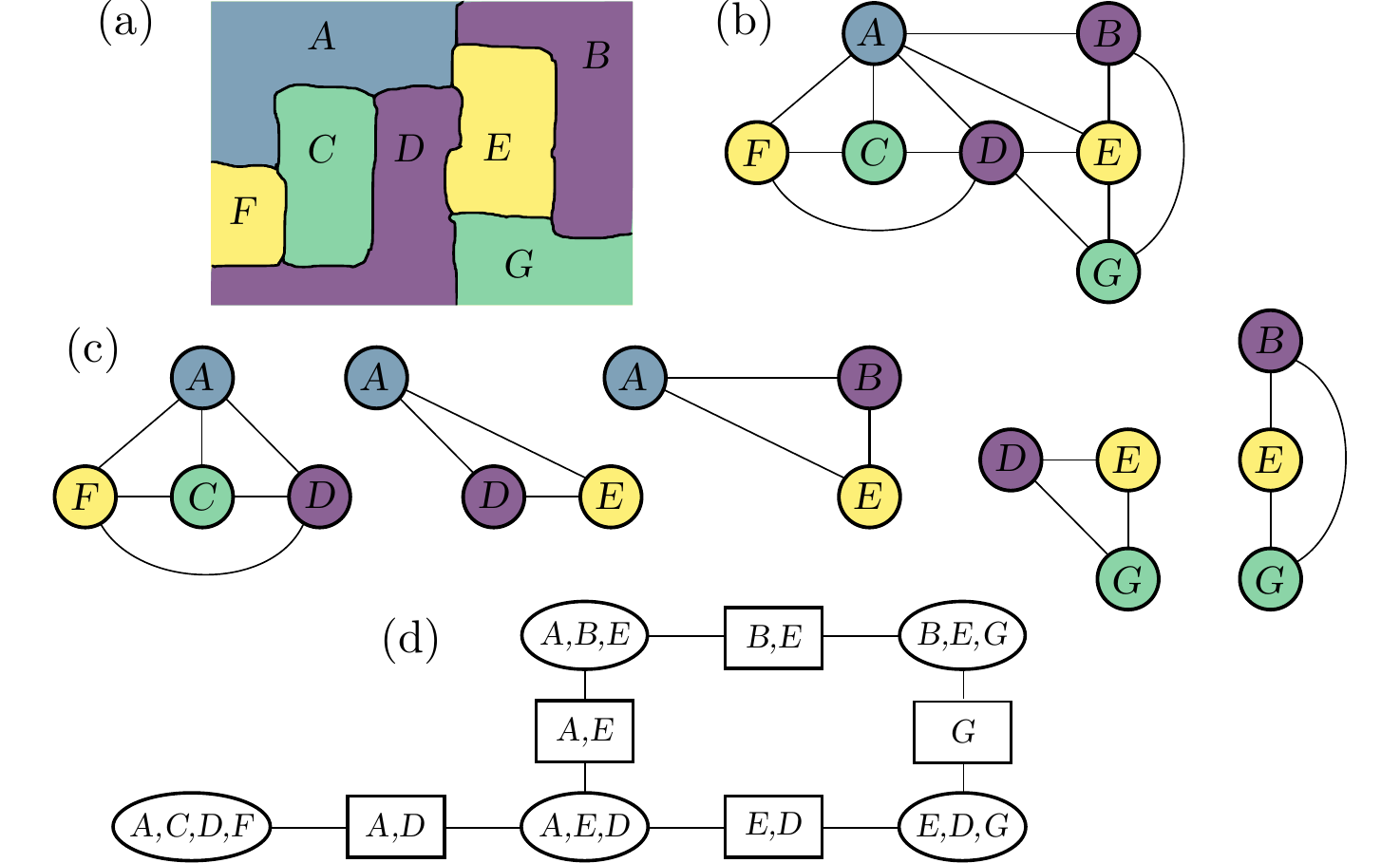}
  \caption{ (a) A four color graph problem containing regions $A$ to
    $G$, with (b) its graph coloring representation, (c) the maximal
    cliques within the graph, and (d) a cluster graph configuration
    for this problem. The ellipses represent the clusters and the
    boxes the sepsets -- see the main text for more detail.
  }\label{fig:fourcolormap}
\end{figure}

A practical example of such a graph coloring is the classical four
color map problem that gave birth to the whole field: a cartographer
is to color the regions of a planar map such that no two adjacent
regions have the same color. To present this problem as graph
coloring, an undirected graph is constructed by representing each
region in the map as a node, and each boundary between two regions as
an edge connecting those two corresponding nodes. Once the problem is
represented in this form, a solution can be approached by any typical
graph coloring algorithm.  An example of this parametrization can be
seen in Figure~\ref{fig:fourcolormap}~(a) and (b); we refer to (c) and
(d) later on.

\subsection{PGMs to represent graph coloring problems} ~\label{sec:gcwithpgms}
PGMs are used as a tool to reason about large-scale probabilistic
systems in a computationally feasible manner. They are known for their
powerful inference over problems with many interdependencies. It is
often useful for problems that are difficult to approach
algorithmically, with graph coloring being a specific example.

In essence a PGM is a compact representation of a probabilistic space
as the product of smaller, conditionally independent, distributions
called factors.  Each factor defines a probabilistic relationship over
the variables within its associated cluster -- a cluster being a set
of random variables. For discrete variables this results in a discrete
probability table over all possible outcomes of these
variables. Instead of explicitly getting the product of these factors
(which typically is not computationally feasible), a PGM connects them
into an appropriate graph structure. Inference is done by passing
messages (or beliefs) over the links in this structure until
convergence is obtained. In combination with the initial factor
distributions, these converged messages can then be used to obtain the
(approximate) posterior marginal distributions over subsets of
variables.

To factorize a graph coloring problem, we first need to parametrize
the problem probabilistically. This is achieved by allowing each node
in the graph to be represented by a discrete random variable $X_i$
that can take on a number of states. For graph coloring these states
are the available labels for the node; e.g. four
colors in the case of the four color map problem.

Now that we have the random variables of our system, and their
domains, we need to capture the relationship between these variables
in order to represent it as factors in our PGM.
For graph coloring, no two adjacent nodes may have the same color,
therefore their associated random variables may not have the same
state. One representation of this system would then be to capture this
relationship using factors with a scope of two variables, each taken
as an adjacent pair of nodes from the coloring graph. Although this is
a full representation of the solution space, there is a trade-off
between accuracy and cluster size (we use size to mean
cardinality)~\cite{mateescu2010join}.

A clique is defined as a set of nodes that are all adjacent to each
other within the graph, and a maximal clique is one that is not fully
contained inside any other clique. To maximize the useful scope of
factors, we prefer to define our factors directly on the maximal
cliques of the graph. (We use the terms clique and cluster more or
less interchangeably.)  We then set the discrete probability tables of
these factors to only allow states where all the variables are
assigned different labels. In the next section we give an example of
this.

After finalizing the factors we can complete the PGM by linking these
factors in a graph structure. There are several valid structure
variants to choose from -- in this paper we specifically focus on
factor graph vs the cluster graph structures. In the resulting graph
structure, linked factors exchange information with each other about
\emph{some}, and not necessarily all, of the random variables they
have in common. These variables are known as the separation set, or
``sepset'' for short, on the particular link of the graph. Whichever
graph structure we choose must satisfy the so-called running
intersection property (RIP)~\cite[p.347]{Koller2009}. This property
stipulates that for all variables in the system, any occurrence of a
particular variable in two distinct clusters should have a unique
(i.e. exactly one) path linking them up via a sequence of sepsets that
all contain that particular variable. Several examples of this are
evident in Figure~\ref{fig:fourcolormap}~(d). In particular note the
absence of the $E$ variable on the sepset between the $\{B,E,G\}$ and
$\{E,D,G\}$ clusters. If this was not so there would have been two
distinct sepset paths containing the variable $E$ between those two
clusters. This would be invalid, broadly because it causes a type of
positive feedback loop.

After establishing both the factors as well as linking them in a graph
structure, we can do inference by using one of several belief
propagation algorithms available.

\subsection{Example: The four color map problem}
We illustrate the above by means of the four color map problem. The
example in Figure~\ref{fig:fourcolormap} can be expressed by the seven
random variables $A$ to $G$, grouped into five maximal cliques as
shown. There will be no clique with more than four variables
(otherwise four colors would not be sufficient, resulting in a
counter-example to the theorem). These maximal cliques are represented
as factors with uniform distributions over their valid
(i.e. non-conflicting) colorings. We do so by assigning either a
possibility or an impossibility to each joint state over the factor's
variables. More specifically we use a non-normalized discrete table and
assign a ``1'' for outcomes where all variables have differing colors,
and a ``0'' for cases with duplicate colors.

For example the factor belief for the clique $\{A,C,D,F\}$ of the
puzzle in Figure~\ref{fig:fourcolormap} is shown in
Table~\ref{tab:probtable}.  These factors are connected into a graph
structure -- such as the cluster graph in
Figure~\ref{fig:fourcolormap}~(d). We can use belief propagation
algorithms on this graph to find posterior beliefs.
\begin{table}[h!]
  \centering
  \begin{tabular}{r c c c c|l l}
    Random variables $\rightarrow$ & $A$ & $C$ & $D$ & $F$ &  &\\
    \hline
    State $\rightarrow$ & 1 & 2 & 3 & 4 & 1 &  $\leftarrow$ $p(A, C, D, F)$\\
    & 1 & 2 & 4 & 3 & 1 &  \ \ non-normalized \\
    & 1 & 3 & 2 & 4 & 1 &  \\
    & 1 & 3 & 4 & 2 & 1 &  \\
    &   &   & $\vdots$  &   &   & \\
    & 4 & 3 & 2 & 1 & 1  & \\\cline{2-6}
    & \multicolumn{4}{c|}{elsewhere} & 0 &
  \end{tabular}
  \caption{A discrete table capturing all possible combinations of
    outcomes for $\{A,C,D,F\}$.}
  \label{tab:probtable}
\end{table}

We successfully tested this concept on various planar maps of size
$100$ up to $8000$ regions. These were generated by first generating
super pixels using the SLIC algorithm~\cite{slic} to serve as the
initially uncolored regions.

We hypothesize that systems configured as described above, utilizing
only binary probabilities, always preserve all possible solutions --
as yet we have found no counterexample to this. (Although this
certainly is not true of loopy graphs making use of non-binary
probabilities). The underlying reason seems to be that a state
considered as possible within a particular factor will always be
retained as such except if a message from a neighboring factor flags
it as impossible. In that case it is of course quite correct that it
should be removed from the spectrum of possibilities.

\begin{figure}[h]
  \begin{center}
    \includegraphics[width=\columnwidth]{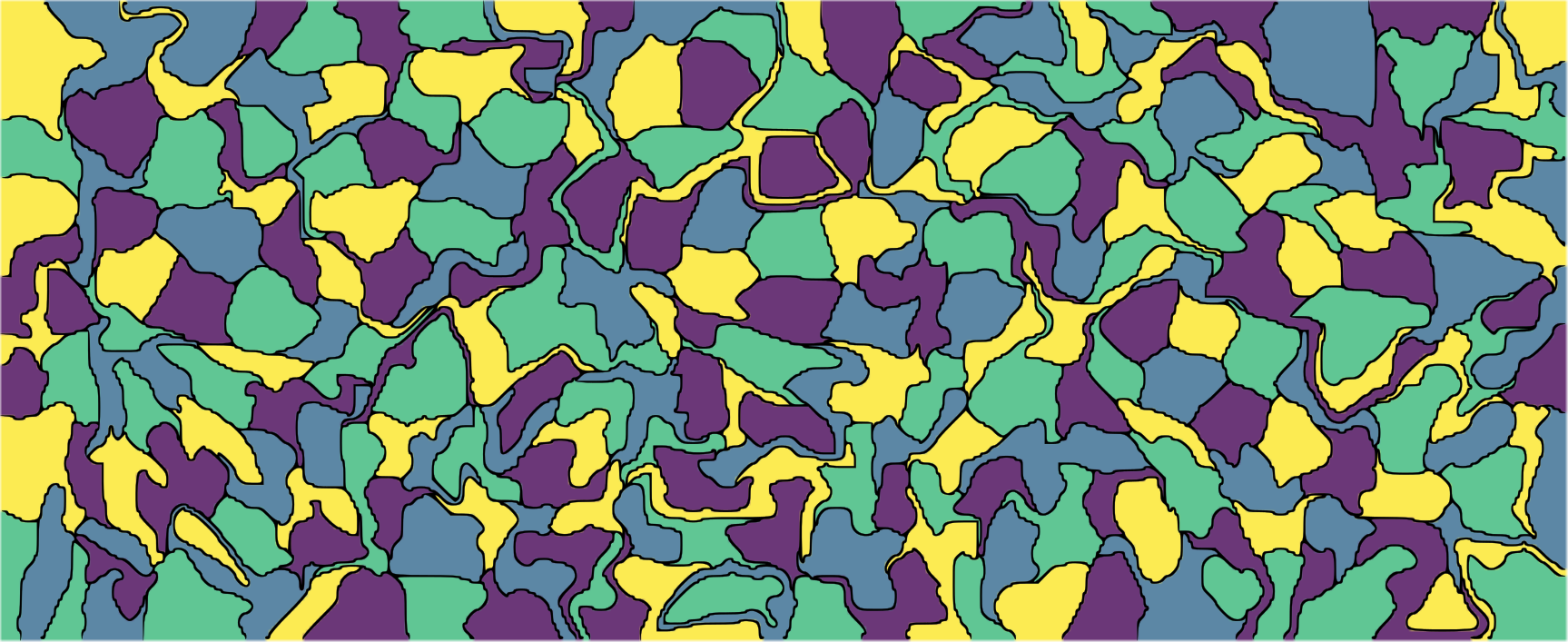}
  \end{center}
  \caption{ A generated planar map resulting from a PGM coloring the
    250 regions into four colors.  }\label{fig:fourcolor}
\end{figure}

However, in this four color map case the space of solutions can in
principle be prohibitively large. We force our PGM to instead find a
particular unique solution, by firstly fixing the colors in the
largest clique, and secondly by very slightly biasing the other factor
probabilities towards initial color preferences. This makes it
possible to pick a particular unique coloring as the most likely
posterior option. An example of a graph of 250 regions can be seen in
Figure~\ref{fig:fourcolor}.

\section{Factor vs cluster graph topologies} \label{sec:topology}
The graph structure of a PGM can make a big difference in the speed
and accuracy of inference convergence. That said, factor graphs are
the predominant structure in literature -- surprisingly so since we
found them to be inferior to a properly structured cluster
graph. Cluster graphs allow for passing multivariate messages between
factors, thereby maintaining some of the inter-variable correlations
already known to the factor. This is in contrast to factor graphs
where information is only passed through univariate messages, thereby
implicitly destroying such correlations.

A search on scholar.google.com (conducted on June 28, 2017) for
articles relating to the use of factor graphs versus cluster graphs in
PGMs returned the following counts:
\begin{itemize}[leftmargin=1.4em]
\item 5590 results for: \textit{probabilistic graphical models "factor graph"},
\item 661 results for: \textit{probabilistic graphical models "cluster graph"}, and
\item 49 results for: \textit{probabilistic graphical models "factor graph" "cluster graph"}.
\end{itemize}
Among the latter 49 publications (excluding four items authored at our
university), no cluster graph constructions are found other than for
Beth\'e / factor graphs, junction trees, and the clustering of Bayes
networks. We speculate that this relative scarcity of cluster graphs
points to the absence of an automatic and generic procedure for
constructing good RIP satisfying cluster graphs.

\subsection{Factor graphs}\label{sec:factorgraphs}
A factor graph, built from clusters $\mathbf{C}_i$, can be expressed
in cluster graph notation as a Beth\'e graph $\mathcal{F}$. For each
available random variable $X_j$, $\mathcal{F}$ contains an additional
cluster $\mathbf{C}_j=\{X_j\}$. Their associated factors are all
uniform (or vacuous) distributions and therefore does not alter the
original product of distributions.  Each cluster containing $X_j$, is
linked to this vacuous cluster $C_j$. This places $C_j$ at the hub of
a star-like topology with all the various $X_j$ subsets radiating
outwards from it. Due to this star-like topology the RIP requirement is
trivially satisfied.

\begin{figure}[h]
  \includegraphics[width=\columnwidth]{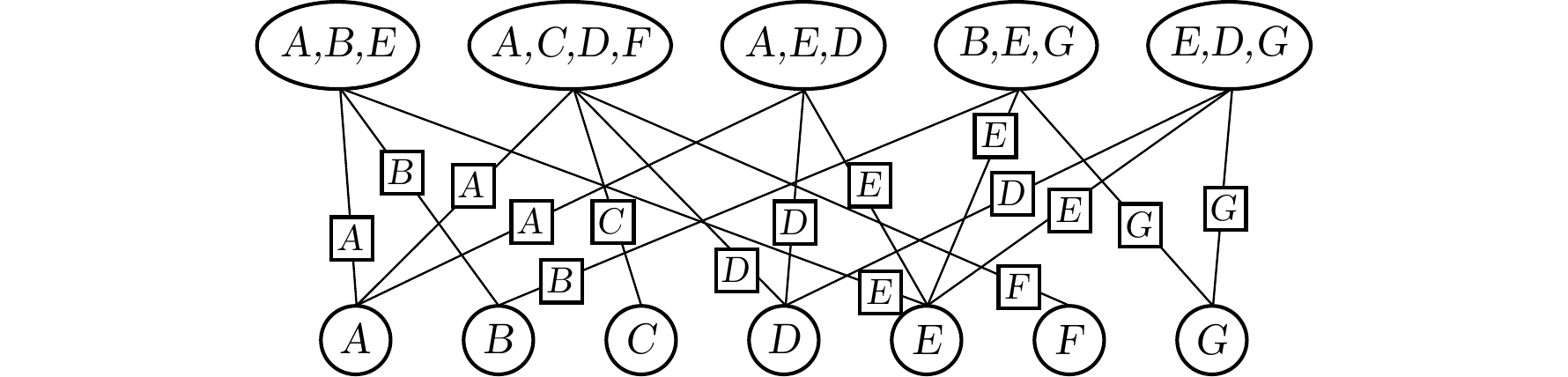}
  \caption{ The Beth\'e factor graph topology applicable to
    Figure~\ref{fig:fourcolormap}. Note the univariate sepsets arranged
    in a star-like topology.  }\label{fig:bethegraph}
\end{figure}

The setup of a factor graph from this definition is straightforward,
the structure is deterministic and the placements of sepsets are well
defined. Figure~\ref{fig:bethegraph} provides the factor graph for the
factors shown in Figure~\ref{fig:fourcolormap}. %[TODO not true at all!!!]

\subsection{Cluster graphs}\label{sec:clustergraphs}
A cluster graph $\mathcal{T}$, built from clusters $\mathbf{C}_i$, is
a non-unique undirected graph, where
\begin{enumerate}%[leftmargin=3em]
\item no cluster is a subset of another cluster, $\mathbf{C}_i
  \nsubseteq \mathbf{C}_j$ for all $i \neq j$,
\item the clusters are used as the nodes,
\item the nodes are connected by non-empty sepsets $\mathbf{S}_{i,j}
  \subseteq \mathbf{C}_i \cap \mathbf{C}_j$,
\item and the sepsets satisfy the running intersection property.
\end{enumerate}
Point (1) is not strictly necessary (see for instance the factor graph
structure), but provides convenient computational savings. It can
always be realized by simply assimilating non-obliging clusters into a
superset cluster via distribution multiplication. Refer to
Figure~\ref{fig:fourcolormap}~(d) for an example of a typical cluster
graph.

Although Koller et al. provides extensive theory on cluster graphs,
they do not provide a general solution for the constructing
thereof~\cite[p.404]{Koller2009}. Indeed, they state that ``the choice
of cluster graph is generally far from obvious, and it can make a
significant difference to the [belief propagation] algorithm.''
Furthermore, the need for such a construction algorithm is made clear
from their experimental evidence, which indicates that faster
convergence and an increase in accuracy can be obtained from better
graph structuring. Therefore, since cluster graph theory is well
established, an efficient and uncomplicated cluster graph construction
algorithm will be useful. We provide the LTRIP algorithm for this
purpose.

\subsection{Cluster graph construction via LTRIP}\label{sec:ltrip}
The LTRIP algorithm is designed to satisfy the running intersection
property for a cluster graph $\mathcal{T}$ by layering the
interconnections for each random variable separately into a tree
structure, and then superimposing these layers to create the combined
sepsets. More precisely, for each random variable ${X}_i$
available in $\mathcal{T}$, all the clusters containing ${X}_i$
are inter-connected into a tree-structure -- this is then the layer
for ${X}_i$. After finalizing all these layers, the sepset
between cluster nodes $\mathbf{C}_i$ and $\mathbf{C}_j$ in
$\mathcal{T}$, is the union of all the individual variable connections
over all these layers.

\begin{figure}
  \includegraphics[width=\columnwidth]{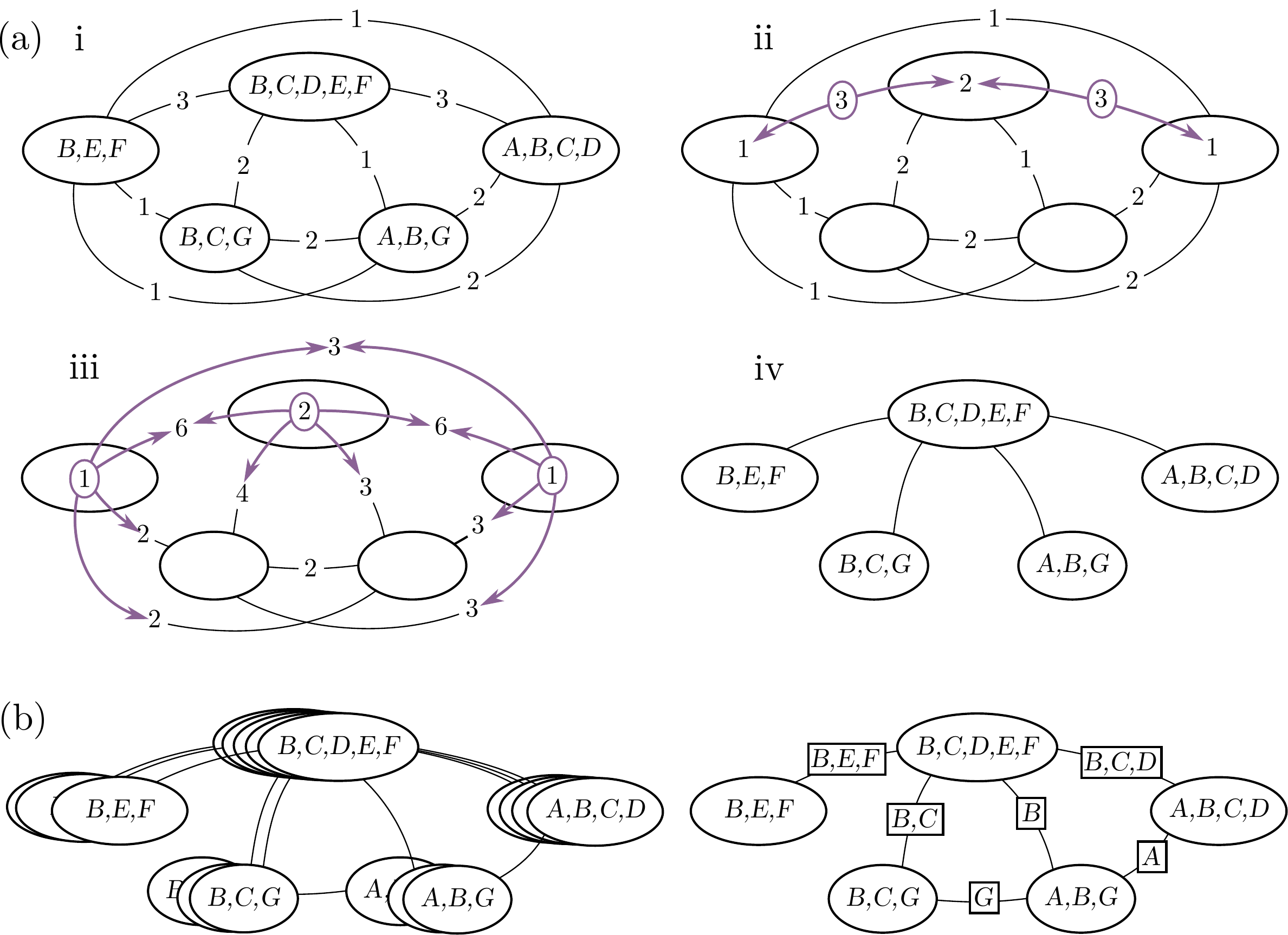}
  \caption{Illustration of constructing a cluster graph via the LTRIP
    procedure. The five clusters to be linked are $\{B,C,D,E,F\}$,
    $\{A,B,C,D\}$, $\{B,E,F\}$, $\{B,C,G\}$ and $\{A,B,G\}$. (a)
    Details the procedure for joining up all clusters containing
    variable $B$ into a tree. In sub-step (i) we set the initial
    connection weights as the number of variables shared by each
    cluster pair. In sub-step (ii) we identify the current maximal
    connection weight to be $m := 3$. In sub-step (iii) we note for
    each cluster how many of its links have maximal weight $m$.  This
    number is added to all its connection weights. This emphasizes
    clusters that are strongly connected to others. In sub-step (iv)
    we use these connection weights to form a maximal spanning tree
    connecting all occurrences of variable $B$.  (b) Similarly
    constructed connection trees for all other variables are
    superimposed to yield the final cluster graph and its
    sepsets.}\label{fig:ltrip}
\end{figure}

While this procedure guarantees satisfying the RIP requirement, there
is still considerable freedom in exactly how the tree-structure on
each separate layer is connected. In this we were guided by the
assumption that it is beneficial to prefer linking clusters with a
high degree of mutual information. We therefore chose to create trees
that maximizes the size of the sepsets between clusters. The full
algorithm is detailed in Algorithm~\ref{alg:ltrip} with an
illustration of the procedure in
Figure~\ref{fig:ltrip}. Note that other (unexplored)
alternatives are possible for the \textsc{connectionWeights} function
in the algorithm.  In particular, it would be interesting to evaluate
information theoretic considerations as criterion.
%and \ref{LTRIP_treeoverlay}.

\def\NoNumber#1{{\def\alglinenumber##1{}\State #1}\addtocounter{ALG@line}{-1}}
\begin{algorithm}
  \caption{\ LTRIP$\,(\,\mathcal{V}\,)$}
  \begin{description}
  \item[\hspace{0.15em}Note:]
  input $\mathcal{V}$ is the set of clusters $\{\mathbf{C}_1, \ldots, \mathbf{C}_N \}$, with subsets already assimilated into their supersets
  %\item[\hspace{0.15em}Output:] Cluster graph $\mathcal{T}$
  \end{description}
  %\vspace{-0.7em}
  %\rule{\columnwidth}{0.45pt}
  \begin{algorithmic}[1]

    \State{// {Empty set of sepsets}}
    \State{$\mathcal{S} := \{\}$ % \ \ // {Empty set of sepsets}
    }
    \For{each random variable $X$ found within $\mathcal{V}$}
    \State{// {This inner loop procedure is illustrated in Figure~\ref{fig:ltrip} (a)}}
    \State $\mathcal{V}_{X}$ := set of clusters in $\mathcal{V}$ containing ${X}$
    %\State{// {Find connection weights between all nodes in $\mathcal{V}_X$}}
    \State{$\mathcal{W}_{X} := $\textsc{connectionWeights}$(\mathcal{V}_X)$}
    \State{// {Add $X$ to the appropriate sepsets}}
    \State $\mathcal{P}_{X}$ := max spanning tree over $\mathcal{V}_{X}$ using weights $\mathcal{W}_{X}$
    \For{each edge $(i,j)$ in $\mathcal{P}_{X}$}
    \If{sepset $\mathbf{S}_{i,j}$ already exists in $\mathcal{S}$}
    \State{$\mathbf{S}_{i,j}\mkern1mu$.\,insert$(X)$}
    \Else
    \State{$\mathcal{S}\mkern1mu$.\,insert$(\mkern2mu \mathbf{S}_{i,j} \mkern-3.5mu = \mkern-3.5mu \{ {X} \} \mkern2mu )$}
    \EndIf
    \EndFor
    \EndFor\vspace{-0.1em}
    \State{$\mathcal{T}:=$ cluster graph of $\mathcal{V}$ connected with sepsets $\mathcal{S}$}\vspace{-0.1em}
    \State{\Return $\mathcal{T}$}\vspace{-0.55em}
        \NoNumber{\hspace{-1.65em}\rule{\columnwidth}{0.42pt}}
    \Function{connectionWeights}{$\mathcal{V}_X$}
            \State $\mathcal{W}_{X}$ := $\{w_{i,j} = \left|\mathbf{C}_i \cap \mathbf{C}_j \right|$ for $ \mathbf{C}_i, \mathbf{C}_j \in \mathcal{} \mathcal{V}_X, \ i \neq j \}$
            \State{// {Emphasize nodes strongly connected to multiple %other  -> cant fit in one line
            nodes}}
            \State $m := \textsc{max}(\mathcal{W}_{X})$
            \For{$i$}
            \State{// {Number of maximal edges on this node}}
            \State{$t_i$ := %$|\{j$ for $w_{i,j} =  m\}|$
                                number of adjacent nodes $j$ for which $w_{i,j} = m$
                }
            \State{// {Add to each edge touching this node}}
            \For{j}
            \State{$w_{i,j}$ += $t_i$ }
            \EndFor
            \EndFor
      \State \Return $\mathcal{W}_{X}$
    \EndFunction
  \end{algorithmic}\label{alg:ltrip}
\end{algorithm}

\section{Modeling Sudoku via PGMs} \label{sec:sudoku}
\begin{figure}[h]
  \includegraphics[width=\columnwidth]{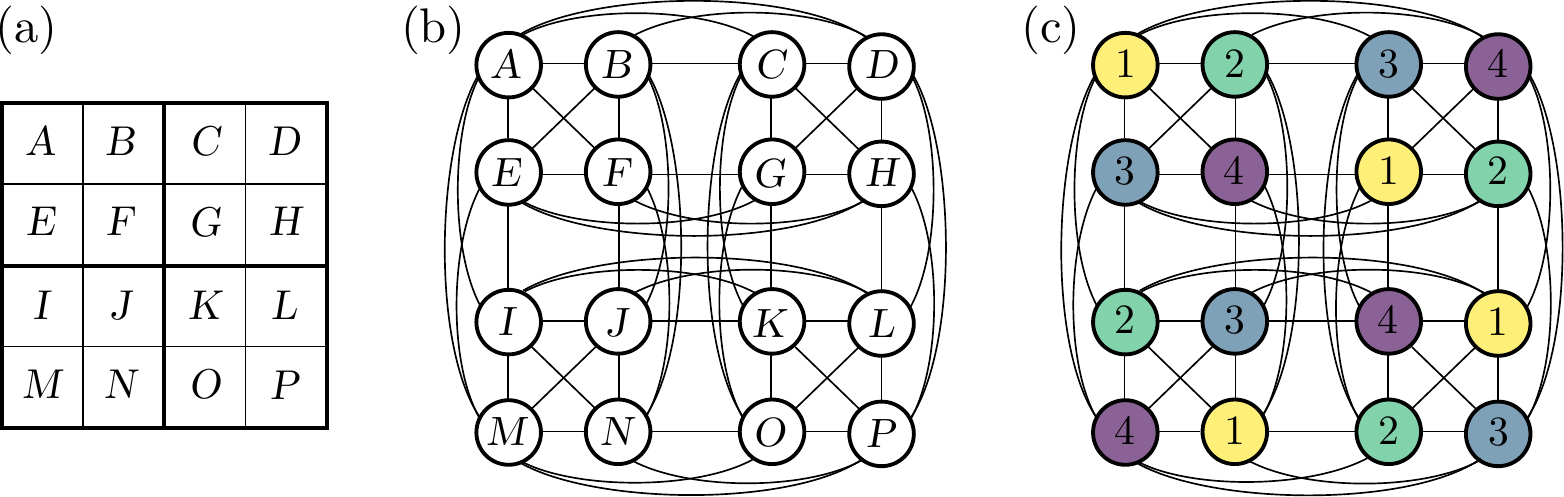}
  \caption{ (a) An example of a $4\times 4$ scaled Sudoku grid, with
    (b) its coloring graph, and (c) a non-unique coloring solution.
  }\label{sudoku}
\end{figure}

The Sudoku puzzle is a well known example of a graph coloring
problem. A player is required to label a $9 \times 9$ grid using the
integers ``1'' to ``9'', such that 27 selected regions have no
repeated entries. These regions are the nine rows, nine columns, and
nine non-overlapping $3 \times 3$ sub-grids of the puzzle. Each label
is to appear exactly once in each region.  If a Sudoku puzzle is
under-constrained, i.e.\ too few of the values are known beforehand,
multiple solutions are possible. A well defined puzzle should have
only a unique solution.  We illustrate these constraints with a
scaled-down $4\times4$ Sudoku (with $2\times 2$ non-overlapping
sub-grids) in Figure~\ref{sudoku}~(a).

%\vspace{-2em}
We use the Sudoku puzzle as a proxy for testing graph coloring via
PGMs, since this is a well known puzzle with many freely available
examples. However, it should be kept in mind that solving Sudoku
puzzles per se is \emph{not} a primary objective of this paper (in
related work not reported on here we have developed a PGM system
capable of easily solving all Sudoku puzzles we have encountered). We
now show how to construct a PGM for a Sudoku puzzle, by following the
same approach as described for the four color map problem.

\subsection{Probabilistic representation}

%Random variables and factors representing their relationships
For the graph coloring and probabilistic representation of the Sudoku
puzzle, each grid entry is taken as a node, and all nodes that are
prohibited from sharing the same label are connected with edges as seen
in Figure~\ref{sudoku}~(b). It is apparent from the graph that each of
the Sudoku's ``no-repeat regions'', is also a maximal clique within
the coloring graph.

The probabilistic representation for the scaled-down $4\times 4$
Sudoku is, therefore, $16$ random variables $A$ to $P$, each
representing a cell within the puzzle. The factors of the system are
set up according to the $12$ cliques present in the coloring graph.
Three examples of these factors, a row constraint, a column constraint
and a sub-grid constraint, are respectively $\{A, B, C, D\}$, $\{A, E,
I, M\}$, and $\{A, B, E, F\}$. The entries for the discrete table of
$\{A, B, C, D\}$ are exactly the same as those of
Table~\ref{tab:probtable}. The proper $9\times 9$ sized Sudoku puzzle used
in our experiments are set up in exactly the same manner than the
scaled down version, but now using $27$ cliques each of size nine.

We should also note that in the case of Sudoku puzzles, some of the
values of the random variables are given beforehand. To integrate this
into the system, we formally ``observe'' that variable. There are
various ways to deal with this, one of which is to purge all the
discrete distribution states not in agreement with the
observations. Following this, the variable can be purged from all
factor scopes altogether.

\subsection{Graph structure for the PGM}

\begin{figure}[h]
  \includegraphics[width=\columnwidth]{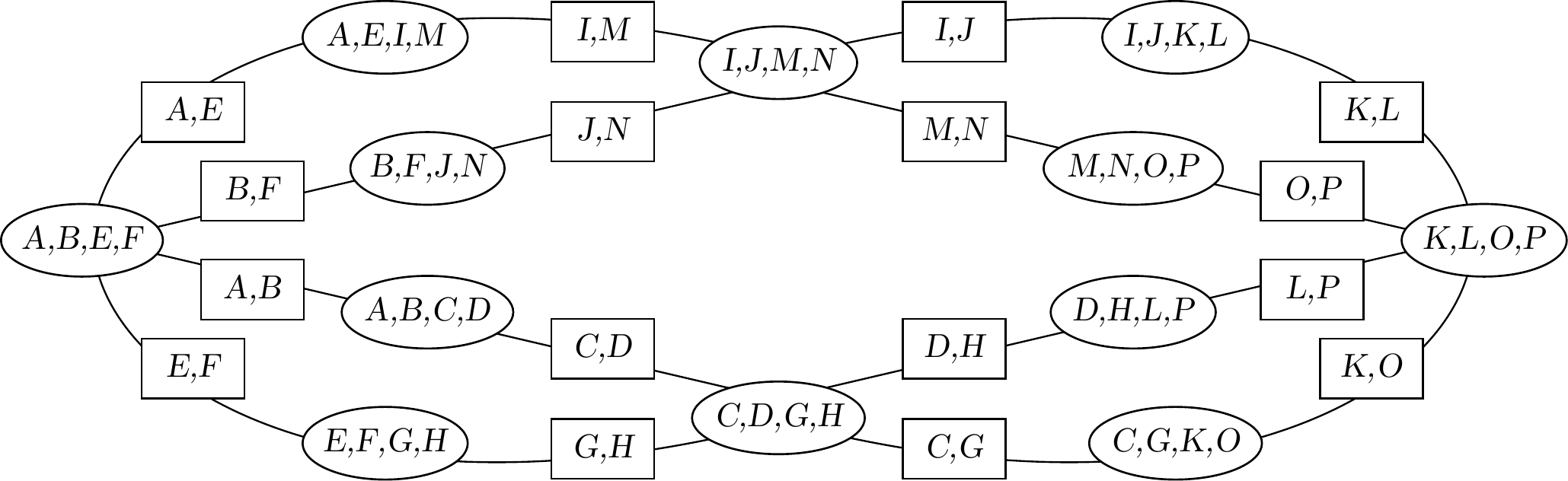}
  \caption{A cluster graph construction for the $4\times 4$ Sudoku clusters.
  }\label{fig:sudoku_construction}
\end{figure}

We have shown how to parametrize the Sudoku puzzle as a coloring
graph, and furthermore, how to parametrize the graph
probabilistically. This captures the relationships between the
variables of the system via discrete probability distributions. The
next step is to link the factors into a graph structure. We outlined
factor graph construction in Section~\ref{sec:factorgraphs}, as well
as cluster graph construction via LTRIP in
Section~\ref{sec:ltrip}. We apply these two
construction methods directly to the Sudoku clusters thereby creating
structures such as the cluster graph of
Figure~\ref{fig:sudoku_construction}.

\subsection{Message passing approach}

For the sake of brevity we do not discuss the detail of belief
propagation techniques here -- this is adequately available from many
resources, including our references. However, for completeness we list
some settings we applied:
\begin{itemize}
\item For the inference procedure we used belief \emph{update}
  procedure, also known as the Lauritzen-Spiegelhalter
  algorithm~\cite{lauritzen1988local},
\item The convergence of the system, as well as the message passing schedule,
  are determined according to Kullback-Leibler divergence between the
  newest and immediately preceding sepset beliefs.
\item Max-normalization and max-marginalization are used in order to
  find the maximum posterior solution over the system.
\item To make efficient use of memory and processing resources all
  discrete distributions support sparse representations.
\end{itemize}

\section{Experimental investigation}\label{sec:experiments}
As stated earlier, factor graphs are the dominant PGM graph structure
encountered in the literature. This seems like a compromise, since
cluster graphs have traits that should enable superior performance. In
this section we investigate the efficiency of cluster graphs compared
to factor graphs by using Sudoku puzzles as test cases.

\subsection{Databases used}
For our experiments, we constructed test examples from two sources,
(a) 50 $9 \times 9 $ Sudoku puzzles ranging in difficulty taken from
Project Euler~\cite{hughes2012problem}, and (b) the ``95 hardest
Sudokus sorted by rating'' taken from Sterten~\cite{sterten}. All
these Sudoku problems are well-defined with a unique solution, and the
results are available for verification.

\subsection{Purpose of experiment}
The goal of our experiments is to investigate both the accuracy as
well as the efficiency of cluster graphs as compared to factor
graphs. Our hypothesis is that properly connected cluster graphs, as
constructed with the LTRIP algorithm, will perform better during
loopy belief propagation than a factor graph constructed with the same
factors.

Mateescu et al.~\cite{mateescu2010join} shows that inference behavior
differs with factor complexities. A graph with large clusters is likely
to be computationally more demanding than a graph with smaller clusters
(when properly constructed from the same system), but the posterior
distribution is likely to be more precise. We therefore want to also
test the performance of cluster graphs compared to factor graphs over
a range of cluster sizes.

\subsection{Design and configuration of the experiment}
Our approach is to set up Sudoku tests with both factor graphs and
cluster graphs using the same initial clusters. With regard to setting
up the PGMs, we follow the construction methodology outlined in
Section~\ref{sec:sudoku}.

In order to generate graphs with smaller cluster sizes, we strike a
balance between clusters of size two using every adjacent pair of
nodes within the coloring graph as described in
Section~\ref{sec:gcwithpgms}, and using the maximal
cliques within the graph, also described in that section. We do so by
generating $M$-sized clusters from an $N$ variables clique (where $M
\leq N$). We split the cliques by sampling all $M$-combination of
variables from the $N$ variable clique, and keeping only a subset of
the samples, such that every pair of adjacent nodes from the clique is
represented at least once within one of the samples.

For experiments using the Project Euler database we construct Sudoku
PGMs with cluster sizes of three, five, seven, and nine variables in
this manner. This results in graphs of $486$, $189$, $108$ and $27$
clusters respectively. We compare the run-time efficiency and solution
accuracy for both factor and cluster graphs constructed from
the same set of clusters.

On the much harder Sterten database PGMs based on cluster sizes
less than nine was very inaccurate. We therefore limit those
experiments to only clusters with size nine.

\subsection{Results and interpretation}
\begin{figure*}
  \vspace{-0.5em}\hspace{-2.5em} %whitespace balancing
  \includegraphics[width=0.915\textwidth]{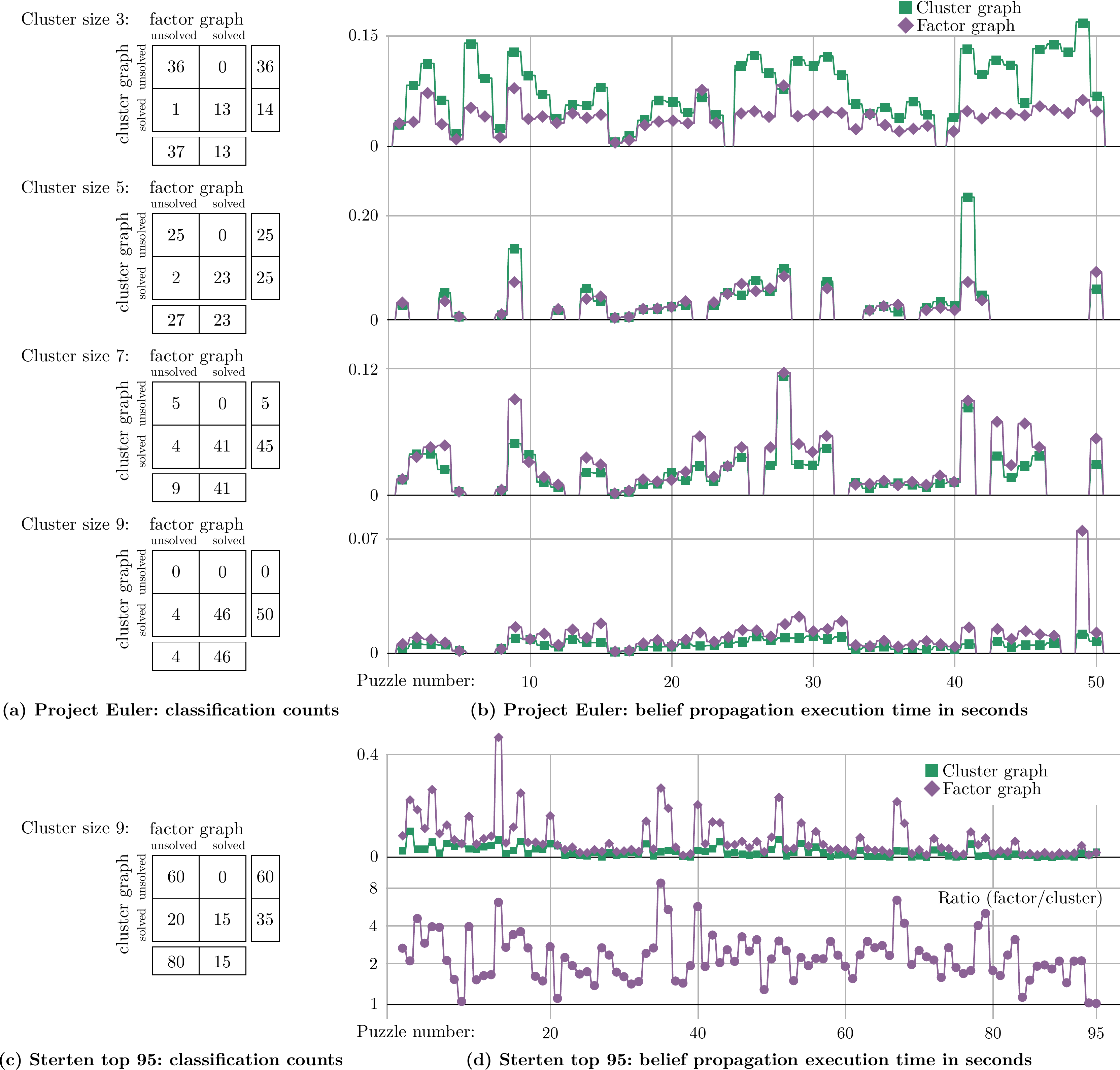}
  \vspace{-0.3em}
  \caption{The results of our test cases.  Note that whenever a
    cluster graph failed to obtain a valid solution, the corresponding
    factor graph also failed.  In (b) we only show results for
    equivalent factor graph and cluster graph posteriors.
  }\label{fig:results}
\end{figure*}

Figure~\ref{fig:results} shows the results we obtained.

Cluster graphs showed superior accuracy for all the available test
cases.  We stress the fact that from our results, whenever a cluster
graph failed to obtain a valid solution the corresponding factor graph
also failed. However, it happened regularly that a cluster graph
succeeded where a factor graph failed, especially so in the more trying
configurations.

In the case of small clusters factor graphs apparently are faster
than cluster graphs. Since cluster graphs built from small clusters
are getting closer to factor graphs in terms of sepset
sizes, this is unexpected. We expected the execution speed to also get
closer to each other in this case.

As the cluster sizes increase, especially so when the problem
domain becomes more difficult, the cluster graphs clearly outperform
the factor graphs in terms of execution speed. Two explanations come
to mind. Firstly, with the larger sepset sizes the cluster graph
needs to marginalize out fewer random variables when passing messages
over that sepset. Since marginalization is one of the expensive
components in message passing, this should result in computational
savings. Secondly, the larger sepset sizes allow factors to pass
richer information to its neighbors. This speeds up the convergence
rate, once again resulting in computational savings.

\section{Future work}
The LTRIP algorithm is shown to produce well constructed graphs. However,
the criteria for building the maximal spanning trees in each layer can
probably benefit from further refinement. In particular we suspect that
taking the mutual information between factors into account might prove useful.

Our graph coloring parametrization managed to solve certain Sudoku
puzzles successfully, as well as assigning colors to the four color
map problem. This is a good starting point for developing more advanced
techniques for solving graph coloring problems.

In this paper we evaluated our cluster graph approach on a limited
set of problems. We hope that the LTRIP algorithm will enhance
the popularity of these problems, as well as other related problems.
This should provide evaluations from a richer set of conditions,
contributing to a better understanding of the merits of this
approach.

\section{Conclusion}
The objective of this study was a) to illustrate how graph coloring
problems can be formulated with PGMs, b) to provide a means for
constructing proper cluster graphs, and c) to compare the performance
of these graphs to the ones prevalent in the current literature.

The main contribution of this paper is certainly LTRIP, our proposed cluster
graph constructing algorithm. %From our experimental results,
The cluster graphs produced by LTRIP show great promise in comparison
to the standard factor graph approach, as demonstrated by our experimental results.

%\begin{acks}
%\end{acks}

\newpage
\bibliographystyle{ACM-Reference-Format}
\balance
\bibliography{graphcoloring}

\end{document}